\documentclass[conference]{IEEEtran}
\IEEEoverridecommandlockouts
\usepackage{cite}
\usepackage{amsmath,amssymb,amsfonts}
\usepackage{algorithmic}
\usepackage{graphicx}
\usepackage{textcomp}
\usepackage{xcolor}
\usepackage{multirow}
\usepackage{subcaption}

\def\BibTeX{{\rm B\kern-.05em{\sc i\kern-.025em b}\kern-.08em
    T\kern-.1667em\lower.7ex\hbox{E}\kern-.125emX}}
\begin{document}

\title{LogGPT: Log Anomaly Detection via GPT}

\author{
\IEEEauthorblockN{Xiao Han}
\IEEEauthorblockA{
\textit{Utah State University}\\
Logan, UT, USA \\
xiao.han@usu.edu}
\and
\IEEEauthorblockN{Shuhan Yuan}
\IEEEauthorblockA{
\textit{Utah State University}\\
Logan, UT, USA \\
Shuhan.Yuan@usu.edu}
\and
\IEEEauthorblockN{Mohamed Trabelsi}
\IEEEauthorblockA{
\textit{Nokia Bell Labs}\\
Murray Hill, NJ, USA \\
mohamed.trabelsi@nokia-\\
bell-labs.com} \\
}

\maketitle

\begin{abstract}
Detecting system anomalies based on log data is important for ensuring the security and reliability of computer systems. Recently, deep learning models have been widely used for log anomaly detection. The core idea is to model the log sequences as natural language and adopt deep sequential models, such as LSTM or Transformer, to encode the normal patterns in log sequences via language modeling. However, there is a gap between language modeling and anomaly detection as the objective of training a sequential model via a language modeling loss is not directly related to anomaly detection. To fill up the gap, we propose LogGPT, a novel framework that employs GPT for log anomaly detection. LogGPT is first trained to predict the next log entry based on the preceding sequence. To further enhance the performance of LogGPT, we propose a novel reinforcement learning strategy to finetune the model specifically for the log anomaly detection task. The experimental results on three datasets show that LogGPT significantly outperforms existing state-of-the-art approaches. 
\end{abstract}


\section{Introduction}
Effectively detecting abnormal events in online computer systems is critical to maintaining the security and reliability of the systems. 
Logs, which are a fundamental component of modern computer systems, serve as a critical source of information for system monitoring, debugging, and security auditing as they record the system status, offering valuable insights into system performance and potential issues. Anomalies in log data often signify system faults, security breaches, or operational failures, making their detection a crucial task \cite{du2017deeplog,guo2021logbert, pang2021deep,le2022log,chalapathy2019deep,landauer2023deep}.

However, the task of anomaly detection in log data is challenging due to the nature of high dimensionality, large volume, and complex structure. 
Machine learning models have been extensively employed for anomaly detection in log data. Traditional models, such as Principal Component Analysis (PCA) \cite{xu2009detecting}, Isolation forest \cite{liu2008isolation}, and one-class Support Vector Machines (OCSVM) \cite{wang2004anomaly} have been widely used. However, these models often require manual feature engineering or assume linear relationships among log entries, which makes them less effective in handling the dynamic nature of log data.

Recently, deep learning models have emerged for log anomaly detection, such as LSTM-based models like DeepLog \cite{du2017deeplog}, LogAnomaly \cite{meng2019loganomaly}, and OC4Seq \cite{wang2021multi}, and BERT-based models like LogBERT \cite{guo2021logbert}. One commonly used strategy is to borrow the idea of language modeling in the natural language processing field to capture the sequential pattern of log data. In this paper, we call this group of log anomaly detection models \textbf{log language model}-based approaches. Particularly, the log language model is first trained to predict the next or masked log entries given the normal sequences. Then, the anomalies can be detected if the observed log entry is not in the top-K list predicted by the log language model. The rationale is that if a log sequence follows normal patterns, the log language model should be able to predict the next or masked log entries. Therefore, when an observed log entry is not in the top-K list predicted by the log language model, it means that the log entry has a low ratio to be in this specific position given the context, indicating the abnormality.

Although empirical studies have demonstrated the effectiveness of leveraging language models for log anomaly detection, the current models still face some limitations. The traditional LSTM-based log language models, such as DeepLog, often fail to fully capture long-term dependencies in log sequences. Therefore, the recently developed models usually adopt the Transformer structure \cite{vaswani2017attention} to model the long log sequences, such as LogBERT \cite{guo2021logbert}. However, the masked log language model adopted in LogBERT may not be able to capture the natural flow in log sequences. 
More importantly, there is a gap between log language modeling and anomaly detection. Technically, the log language model is usually trained to correctly predict the next log entry, while the current log anomaly detection models label the anomalies if the observed log entry is not in the Top-K list predicted by the log language model. In other words, there is a gap in the objective between the training phase and the testing phase for log anomaly detection.



Inspired by the training strategy for large language models, to fill up the gap, we introduce LogGPT, a novel framework for log anomaly detection that leverages the Generative Pre-trained Transformer (GPT) model. LogGPT still harnesses the power of generative log language models to capture the intricate patterns and dependencies in log data. Specifically, LogGPT is pre-trained to predict the next log entry given the preceding sequence (prompt). More importantly, we further fine-tune LogGPT via reinforcement learning. Specifically, LogGPT employs a novel reward mechanism based on whether the observed log entry is within the Top-K predicted log entries from the log language model. If the observed log entry is found within the Top-K predictions, LogGPT will receive a positive reward; otherwise, it will receive a negative reward. Reinforced by this reward signal, we expect that for the normal sequences, LogGPT can ensure the log entry is within the Top-K predictions.



The contributions of this paper are threefold. First, we propose LogGPT, a novel framework for anomaly detection in log data, which utilizes the generative log language model to capture the patterns of normal log sequences by training to predict the next log key given the previous sequence. This novel approach effectively addresses the limitations of both traditional machine learning models and deep learning models like DeepLog \cite{du2017deeplog} and LogBERT \cite{guo2021logbert}, providing a more robust and effective solution for log anomaly detection. Second, we introduce a Top-K reward metric specifically designed for fine-tuning the log language model for anomaly detection. This reward metric gives a positive reward if the actual log key is in the Top-K predictions, and a negative reward otherwise, thereby guiding the model to focus on the most relevant parts of the log sequence and enhancing the accuracy of anomaly detection. Third, we conduct extensive experiments to validate the effectiveness of LogGPT in detecting anomalies in log data. Experimental results demonstrate that LogGPT outperforms state-of-the-art methods, underscoring its potential as a powerful tool for anomaly detection in log data.

\section{Related Work}

Log anomaly detection, a critical task for ensuring system security and reliability, has received extensive research. The methods for log anomaly detection can be broadly categorized into two phases: traditional machine learning models and deep learning models.

In the early phase, traditional machine-learning models were the primary tools for log anomaly detection. Models such as Principal Component Analysis (PCA) \cite{xu2009detecting}, Isolation forest \cite{liu2008isolation}, and one-class Support Vector Machines (OCSVM) \cite{wang2004anomaly} were commonly used. Although these models are capable of identifying outliers in the log data, these models have several limitations. First, the traditional machine learning models usually require manual feature engineering, which is labor-intensive and might not capture the complex patterns in log data. Furthermore, these models struggle with capturing complex patterns in log sequences.


The advanced deep learning models have significantly improved the performance of log anomaly detection. In particular, Long Short-Term Memory Networks (LSTMs), known for their ability to model sequential data, have proven to be effective for log anomaly detection, such as DeepLog \cite{du2017deeplog} and LogAnomaly \cite{meng2019loganomaly}. DeepLog functions by predicting the next log key based on the preceding sequence, identifying anomalies when the actual next log key significantly deviates from the prediction. On the other hand, LogAnomaly models a log stream as a natural language sequence and develops template2vec to extract the semantic information hidden in log templates. Therefore, LogAnomaly can detect both sequential and quantitative log anomalies simultaneously. 
However, these models come with their own set of limitations. A primary challenge with LSTM is that this type of recurrent architecture struggles to encode very long or complex sequences due to its relatively simple structure. 
This issue is particularly pronounced in log anomaly detection, where the sequences can be quite long and complex.

To address the limitations of LSTM-based models, researchers have turned to the use of Transformer \cite{devlin2018bert}, which is a more powerful model to capture the long-term dependencies in the sequences, such as LogBERT \cite{guo2021logbert} or CAT \cite{zhang2022cat}. LogBERT is a self-supervised framework that learns the patterns of normal log sequences based on BERT \cite{devlin2018bert}. Specifically, LogBERT takes normal log sequences with random masks as inputs and is trained to predict the randomly masked log entries. After training, LogBERT can encode the patterns of normal log sequences. One limitation is that the masked log language model may not always capture the natural flow of log sequences in some contexts. Moreover, the performance of LogBERT is sensitive to the mask ratio, a hyperparameter controlling how many tokens will be replaced with MASK tokens during both the training and testing phases. 
In this work, we propose LogGPT, which leverages the GPT model to learn patterns in normal log sequences by predicting the next log entries in a sequence, and further proposes a novel reinforcement learning mechanism to enhance the performance for anomaly detection.

\section{Preliminary}
\begin{figure}
\centering
    \includegraphics[width=0.48\textwidth]{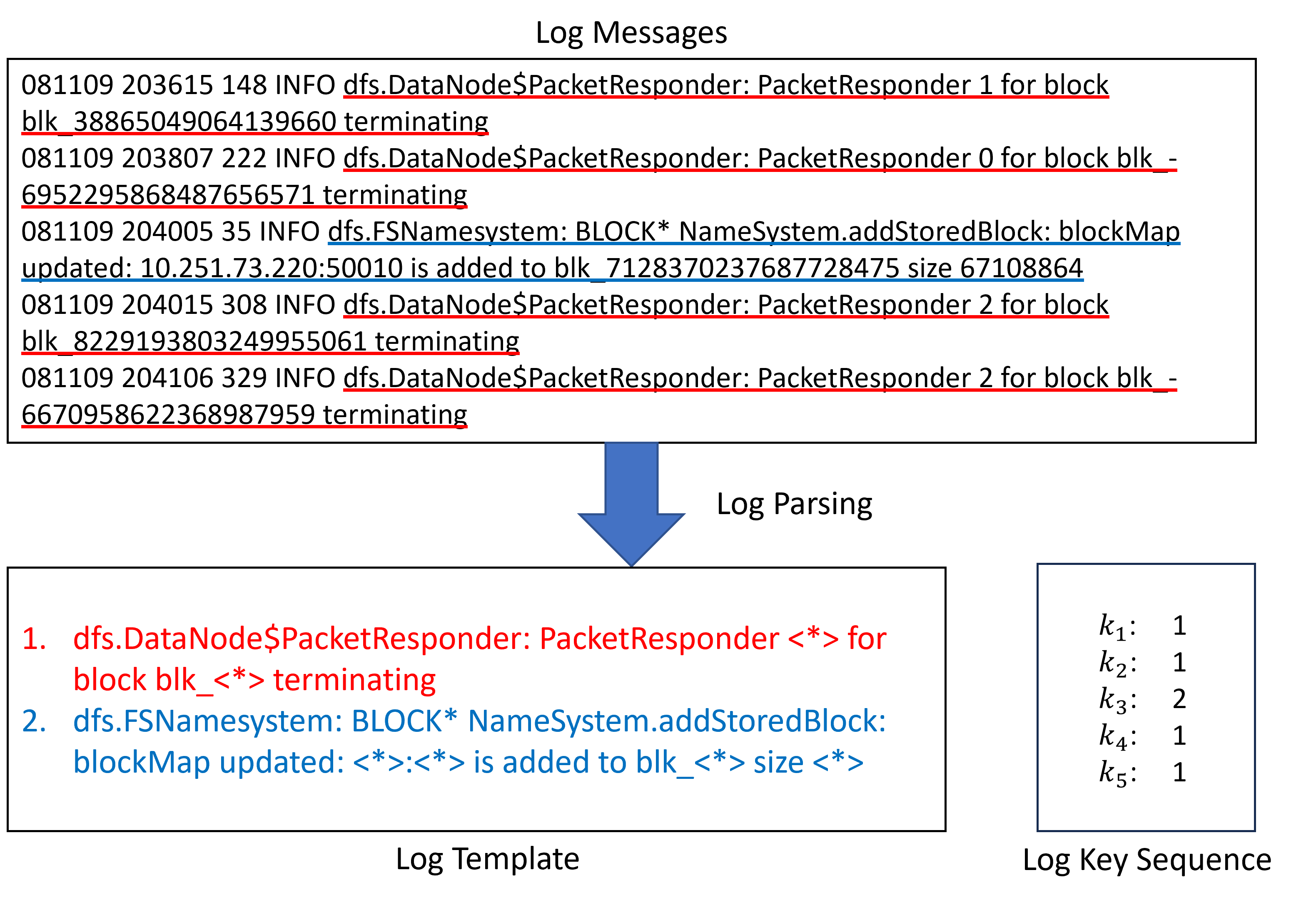}
    \caption{Log key extraction from HDFS dataset messages via Log Parser. The message with a red/blue underscore indicates the detailed computational event for each log key separately.}
\label{fig:parser}  
\end{figure}
In this section, we provide a detailed overview of two key components for log anomaly detection, log sequence preprocessing and log language model.

\subsection{Log Sequence Preprocessing}
The first step of log anomaly detection is to preprocess the log messages because it is hard to capture the sequential pattern from the raw text-based log messages. The major line of research in log anomaly detection is to first adopt a log parser, such as Drain \cite{he2017drain}, to extract the template from the log messages, as shown in Figure \ref{fig:parser}. Each template usually indicates one type of log message, called a log key. 

After getting the log keys, the sequence of raw log messages can be transformed into a sequence of log keys. In this case, the log keys are similar to the vocabulary in natural language, while the sequence is like a sentence consisting of a sequence of log keys. Therefore, a language model can be leveraged to model the log sequences. 

Formally, after preprocessing, the log messages with the same template are represented by a log key  $k\in\mathcal{K}$, where $\mathcal{K}$ indicates the set of log keys extracted from the log messages. Then, a log sequence is organized as ordered log keys, denoted as $S=\{k_1,...,k_t,...,k_T\}$, where $T$ indicates the length of the log sequence.


\subsection{Log Language Model}
We use DeepLog \cite{du2017deeplog} to illustrate the concept of the log language model. DeepLog leverages Long Short-Term Memory networks (LSTMs) for log language modeling. The primary objective of DeepLog is to learn a probabilistic model of normal execution from log data and then detect anomalies as significant deviations from normal patterns.

DeepLog is trained on $\mathcal{D}=\{S^i\}_{i=1}^N$ consisting of normal log sequences. 
The LSTM network in DeepLog is trained to predict the next log key in a sequence based on the preceding sequence. Formally, given a sequence of log keys $S_{1:T} = \{k_1, ..., k_t, ..., k_T\}$, where $k_t$ indicates the log key at the $t$-th position. DeepLog trains an LSTM to model the conditional probability $p(k_{t+m+1} | S_{t:t+m})$ for $t = 1, 2, ..., T-m-1$, where $m$ indicates the window size.
Particularly, DeepLog adopts a sliding window with size $m$ to split the sequences into a set of small windows and predict the next log key given the previous $m$ log keys.
The LSTM is trained to maximize the likelihood of the next log key given the preceding sequence, which can be formulated as the following objective function:
\begin{equation}
    \mathcal{L}(\theta) = -\frac{1}{N}\sum_{i=1}^N \sum_{t=1}^{T-m-1} \log p(k_{t+m+1}^i | S_{t:t+m}^i),
\end{equation}
where $\theta$ denotes the parameters of LSTM.

During the anomaly detection phase, given a new sequence, DeepLog still splits the sequences into small windows and employs the trained LSTM model to predict the next log key. The LSTM model predicts a probability distribution over all possible log keys in $\mathcal{K}$, ranking them based on their likelihood of being the next key in the sequence. Then, an abnormal sequence will be labeled as abnormal if the observed log key does not appear in the Top-K prediction list multiple times across all sliding windows in that sequence. 

The concept of Top-K predictions is introduced to account for the inherent uncertainty and variability in log sequences. Even in normal operations, there can be multiple valid ``next'' log keys as the systems usually have multiple normal patterns. Therefore, during the anomaly detection phase, instead of predicting a single `most likely' next log key, the model identifies the Top-K most probable next log keys. As long as the observed log key is in the Top-K list, we could consider the sequence normal.

The value of K, a tunable hyperparameter, determines the strictness of the model for anomaly detection. A smaller K results in a stricter model that allows fewer possibilities for the next log key, usually leading to high recall and low precision, while a larger K results in a more flexible model that considers a broader range of log keys as normal, usually resulting in high precision and low recall.

\section{LogGPT}
In this section, we introduce LogGPT, a novel log anomaly detection model based on GPT. Similar to DeepLog, LogGPT detects the log anomaly by examining whether the observed log key is in the Top-K prediction list. Because GPT is a more powerful structure compared to LSTM used by DeepLog, LogGPT does not need to further split the sequence into multiple small windows. Instead, LogGPT is trained to predict the next log key given the previous sequence, which intrinsically can capture the long-term dependence of log sequences.
Moreover, besides leveraging the powerful GPT structure, we also propose a novel reinforcement learning strategy to further improve the performance of log anomaly detection.




\begin{figure*}[h]
    \centering
    \begin{subfigure}[b]{0.34\textwidth}
        \centering
        \includegraphics[width=\textwidth]{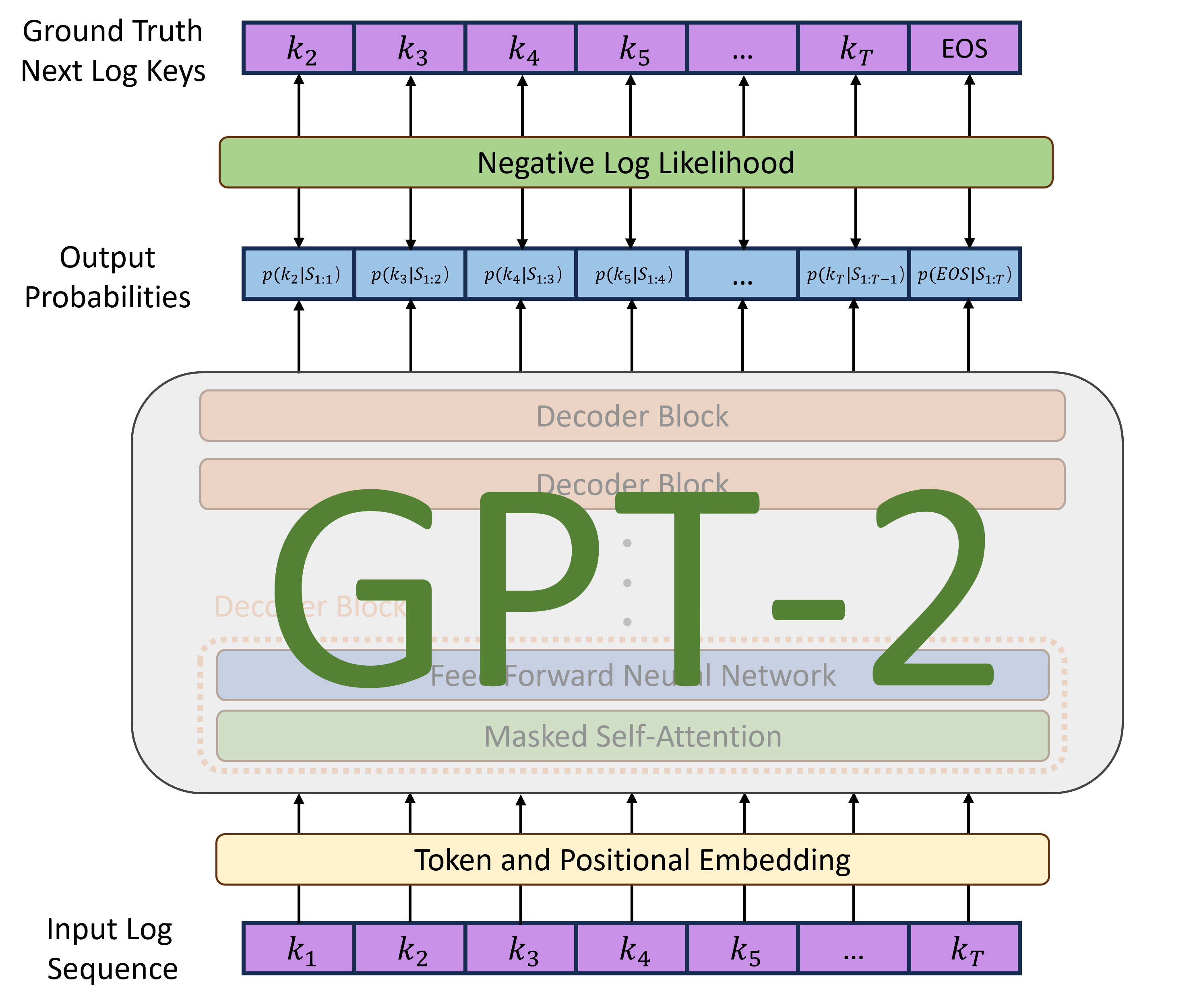}
        \caption{Pre-training}
        \label{fig:pretraining}
    \end{subfigure}
    \hfill
    \begin{subfigure}[b]{0.60\textwidth}
        \centering
        \includegraphics[width=\textwidth]{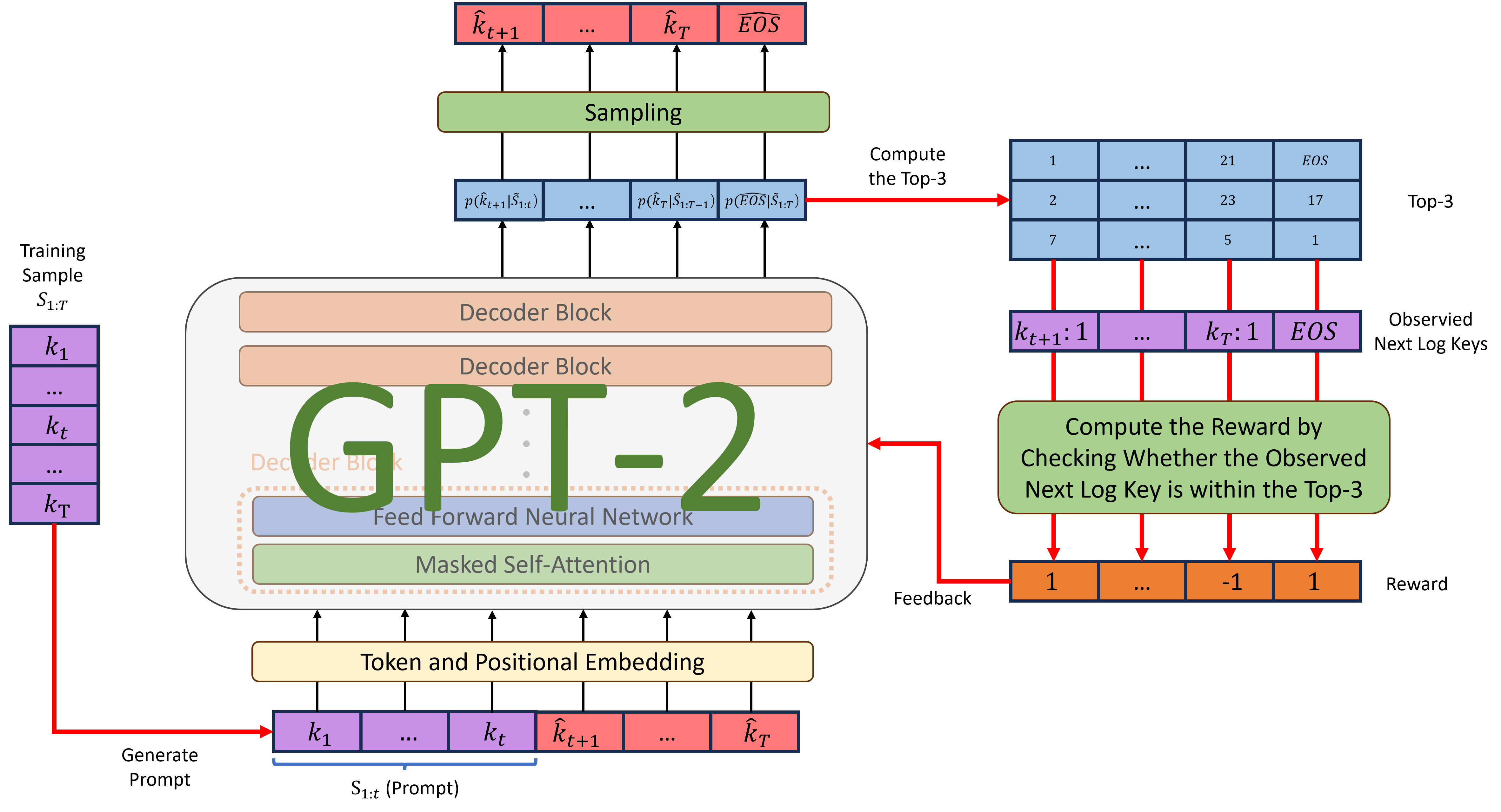}
        \caption{Fine-tuning}
        \label{fig:finetuning}
    \end{subfigure}
    \caption{Framework of LogGPT.}
    \label{fig:framework}
\end{figure*}

The design of LogGPT is inspired by the training process of large language models, where the training process consists of two primary stages: pre-training and fine-tuning, as shown in Figure \ref{fig:framework}.

In the pre-training stage (Figure \ref{fig:pretraining}), a generative log language model $f_{\theta}(\cdot)$ is trained on a corpus of normal log sequences $\mathcal{D}$, which allows the model to learn the underlying patterns and structures of normal system behavior. After pre-training, LogGPT is capable of generating log sequences based on a given part of the log sequences.

The fine-tuning stage (Figure \ref{fig:finetuning}) is designed to further refine the model's ability to distinguish between normal and abnormal log sequences. In this stage, we employ reinforcement learning techniques to finetune the pre-trained LogGPT. Borrowing the terminology from the large language model, we define a set of prompts $\mathcal{P} = \{S^i_{1:t}\}_{i=1}^N$, where $S_{1:t}^i \subseteq S_{1:T}^i$ and $S_{1:T}^i \in \mathcal{D}$. These prompts are fed into the LogGPT to generate the following sequence $\hat{S}_{t:T}^i$ step by step. We propose a novel reward, called the Top-K metric, to fine-tune LogGPT for anomaly detection.


\subsection{Generative Log Language Model}
LogGPT utilizes GPT-2 \cite{radford2019language} for modeling the log sequences, which is based on Transformer decoder \cite{vaswani2017attention} that utilizes a self-attention mechanism to capture dependencies between log keys in the log sequence. LogGPT is trained to predict the next log key given the preceding log keys. The objective function for pretraining the LogGPT is defined as follows:
\begin{equation}
\label{eq:lm}
    \mathcal{L}(\theta) = -\frac{1}{N}\sum_{i=1}^N\sum_{t=1}^{T-1}\log p(k_{t+1}^i | S_{1:t}^i),
\end{equation}
where $\theta$ denotes the parameters of LogGPT, $N$ is the number of log sequences and $T$ is the length of each sequence, $p(k_{t+1}^i | S_{1:t}^i)$ indicates the probability of log key at the $t+1$-th position predicted by LogGPT given the sequence $S_{1:t}^i$.

Specifically, to derive $p(k_{t+1}^i | S_{1:t}^i)$, the structure of LogGPT can be defined as:

\begin{subequations}
\label{eq:next_key}
\begin{align}
    \mathbf{h}_{t}^i &= \textsf{Transformer\_Decoder}(S_{1:t}^i) \label{eq:rep} \\
    p(k_{t+1}^i | S_{1:t}^i) &= \textsf{Softmax}(\mathbf{h}_{t}^i \mathbf{W}), \label{eq:softmax}
\end{align}
\end{subequations}
where $\mathbf{h}_{t}^i \in \mathbb{R}^d$ indicates the hidden representation derived from the Transformer decoder \cite{vaswani2017attention, radford2019language}, and $\mathbf{W} \in \mathbb{R}^{d \times |\mathcal{K}|}$ is the parameter of the language model head that maps the hidden representation to a probability distribution of all log keys in $\mathcal{K}$.




By training the model to predict the next log key in normal log sequences, LogGPT encodes the normal system behavior. After pre-training, GPT-2 is capable of generating a log sequence $\hat{S}_{t+1:T}^i=\{\hat{k}_{t+1}^i,..., \hat{k}_{T}^i\}$ based on a given part of the log sequence $S_{1:t}^i$. This capability is crucial for the subsequent fine-tuning stage, where the model is further refined to distinguish between normal and anomalous log sequences.

\subsection{Reinforcement Learning for Log Anomaly Detection}
In the context of LogGPT, we employ reinforcement learning to fine-tune the pre-trained GPT-2 model for the task of log anomaly detection. The reinforcement learning paradigm is particularly suitable for our task as it allows the model to learn from its predictions and adjust its behavior based on the feedback received, thereby enhancing its ability to detect anomalies.
In the context of our framework, we define the following elements.

\noindent\textbf{State:}  The state, denoted as $\tilde{S}_{1:t}^i = S_{1:t}^i$, is initially defined as the given part of a log sequence. As the model generates the log sequence $\hat{S}_{t+1:T}^i$ based on the given part, the state evolves dynamically. Specifically, for each step $j$ where $t+1 \leq j \leq T-1$, the state $\tilde{S}_{1:j}^i$ becomes the concatenation of the given part of the log sequence $S_{1:t}^i$ and the generated part of the log sequence $\hat{S}_{t+1:j}^i$, denoted as $\tilde{S}_{1:j}^i= \{S_{1:t}^i, \hat{S}_{t+1:j}^i\}$. The sequence $\tilde{S}_{1:j}^i$ is further transformed to a hidden representation $\mathbf{\tilde{h}}_j^i$ by the Transformer decoder shown in Equation \ref{eq:rep}.

\noindent\textbf{Action:} An action is defined as sampling a log key from the K log keys with the highest probabilities predicted by LogGPT, denoted as $a_{j+1}^i \sim \text{Top-K}(p(\hat{k}_{j+1}^i|\tilde{S}_{1:j}^i))$.  

\noindent\textbf{Policy:} A policy takes the form of LogGPT and is defined by its parameters. Specifically, given the current part of the sequence until the $j$-th position, the policy outputs a probability distribution over the action space, represented as $\pi_\theta(a_{j+1}^i | \mathbf{\tilde{h}}_j^i)$, where $\theta$ indicates the parameters of LogGPT.

\noindent\textbf{Reward:} The reward function provides feedback to the policy based on the quality of its actions. We propose a novel reward function to evaluate the predicted log key for anomaly detection, called the Top-K metric. 

At each step, the Top-K metric checks whether the observed next log key is within the Top-K predicted log keys. If this is the case, the model receives a reward of 1; otherwise, it receives a reward of -1. Given a part of log sequence $S_{1:t}^i$, after an action is taken, the reward function is formulated as:
\begin{equation}
    r_{j+1} = \begin{cases}
        1, & \text{if } k_{j+1}^i \in \text{Top-K}(p(\hat{k}_{j+1}^i|\tilde{S}_{1:j}^i)) \\
        -1, & \text{if } k_{j+1}^i \notin \text{Top-K}(p(\hat{k}_{j+1}^i|\tilde{S}_{1:j}^i))
        \end{cases}.
\end{equation}
Here, $k_{j+1}^i$ refers to the actual next log key, and $p(\hat{k}_{j+1}^i|\tilde{S}_{1:j}^i)$ denotes the probability distribution predicted by LogGPT over the action space given the current state. 

The Top-K metric promotes better generalization and robustness of LogGPT in anomaly detection. By encouraging the model to predict a set of likely next log keys rather than a single most likely log key, the Top-K metric helps LogGPT learn a more nuanced representation of the normal log patterns. This approach recognizes that log data may contain inherent variability even for the normal log sequences, and a broader range of acceptable candidates can still reflect normal system behavior. The Top-K metric, therefore, enhances the precision of anomaly detection by aligning the model's predictions with the complex nature of log data.



\noindent\subsection{Policy Update}
We adopt Proximal Policy Optimization (PPO) \cite{schulman2017proximal} for the policy update. PPO is a type of policy gradient method that optimizes the policy directly by maximizing the expected reward and can further maintain the stability of the learning process and prevent harmful updates. 
The objective function of PPO is defined as follows:
\begin{equation}
\label{eq:ppo}
    \begin{split}
        J(\theta) = \mathbb{E}_{\pi_\theta}\left[\sum_{i=1}^N \sum_{j=t}^{T-1} \frac{\pi_\theta(a_{j+1}^i|\mathbf{h}_{j}^i)}{\pi_{\theta_{\text{old}}}(a_{j+1}^i|\mathbf{h}_{j}^i)} r_{j+1}\right],
    \end{split}
\end{equation}
where $\pi_{\theta}$ is the new policy, $\pi_{\theta_{\text{old}}}$ is the old policy, and $r_{j+1}$ is the reward for an action.

The policy $\pi_{\theta}$ is updated by performing gradient ascent on the objective function $J(\theta)$:
\begin{equation}
\label{eq:update}
    \theta \leftarrow \theta + \alpha \nabla_{\theta} J(\theta),
\end{equation}
where $\alpha$ is the learning rate.

The policy update process is repeated for a number of iterations until the policy converges or a maximum number of iterations is reached. 
The Top-K metric encourages the model to recognize the inherent variability in normal log data by rewarding predictions that include the actual next log key within a broader set.

\subsection{Anomaly Detection}
After fine-tuning, LogGPT is deployed to detect abnormal log sequences. 
Given a new log sequence $S_{1:T}$, LogGPT iteratively predicts the next log key $k_{t+1}$ given the preceding subsequence $S_{1:t}$ for $1 \leq t \leq T-1$.

For each predicted log key, the model generates a set of Top-K predicted log keys. This set represents the K most likely log keys at the current position. The actual next log key is then compared to this set. As long as one actual log key is not in the set of Top-K predicted log keys, the whole log sequence will be flagged as anomalous.

\section{Experiments}
\subsection{Experimental Setup}
\begin{table}[]
\caption{Statistics of the Datasets. The number in the parentheses indicates the unique log keys in the training set.}
\label{tab:data-statistics}
\resizebox{0.48\textwidth}{!}{%
\begin{tabular}{|c|c|c|c|c|cc|}
\hline
\multirow{2}{*}{Dataset} & \multirow{2}{*}{\begin{tabular}[c]{@{}c@{}}\# of Unique\\ Log Keys\end{tabular}} & \multirow{2}{*}{\begin{tabular}[c]{@{}c@{}}\# of Log \\ Sequences\end{tabular}} & \multirow{2}{*}{\begin{tabular}[c]{@{}c@{}}Avg. Seq.\\ Length\end{tabular}} & \multirow{2}{*}{\begin{tabular}[c]{@{}c@{}}Training\\ Data\end{tabular}} & \multicolumn{2}{c|}{Testing Data}        \\ \cline{6-7} 
                         &                                                                                  &                                                                                 &                                                                             &                                                                          & \multicolumn{1}{c|}{Normal}  & Anomalous \\ \hline
HDFS                     & 48 (15)                                                                          & 575,061                                                                         & 19                                                                          & 5,000                                                                    & \multicolumn{1}{c|}{553,223} & 16,838    \\ \hline
BGL                      & 396 (160)                                                                        & 36,927                                                                          & 58                                                                          & 5,000                                                                    & \multicolumn{1}{c|}{28,631}  & 3,296     \\ \hline
Thunderbird              & 7,703 (904)                                                                       & 112,959                                                                         & 166                                                                         & 5,000                                                                    & \multicolumn{1}{c|}{67,039}  & 40,920    \\ \hline
\end{tabular}}
\end{table}
\noindent\textbf{Datasets.} We evaluate LogGPT on three log datasets, namely HDFS, BGL, and Thunderbird. Table \ref{tab:data-statistics} shows the statistics of three datasets. For all the datasets, we randomly select 5000 normal log sequences as the training dataset.
\begin{itemize}
    \item HDFS (Hadoop Distributed File System) \cite{xu2009detecting}: This dataset is derived from Hadoop-based map-reduce jobs that were run on Amazon EC2 nodes. The anomalies within this dataset are identified through a manual labeling process based on a set of predefined rules.  The log sequences are constructed based on the session ID present in each log message, resulting in an average sequence length of 19.
    The HDFS dataset consists of 575,061 log sequences, out of which 16,838 have been labeled as anomalous. 
    \item BGL (BlueGene/L Supercomputer System) \cite{oliner2007supercomputers}: The BGL dataset originates from a BlueGene/L supercomputer system, located at the Lawrence Livermore National Labs (LLNL). It includes both alert and non-alert messages, with the alert messages being treated as anomalies. Log sequences are formed using a time sliding window of 1 minute, yielding an average sequence length of 58.
    The BGL dataset contains 36,927 log sequences, with 3,296 of them classified as anomalous. 
    \item Thunderbird \cite{oliner2007supercomputers}: This dataset is collected from another supercomputer system. The dataset used in this study comprises the first 20,000,000 log messages from the original Thunderbird dataset that compose 112,959 log sequences, with 40,920 of them marked as anomalous. Log sequences are created using a time sliding window of 1 minute, leading to an average sequence length of 166.
\end{itemize}

\noindent\textbf{Baselines.} We compare LogGPT with a variety of baseline methods, consisting of both traditional machine learning models and deep learning models:
\begin{itemize}
    \item PCA (Principal Component Analysis) \cite{xu2009largescale}: This technique constructs a counting matrix based on the frequency of log key sequences. It then reduces this matrix into a lower-dimensional space to identify anomalies.
    \item iForest (Isolation Forest) \cite{liu2008isolation}: iForest is an unsupervised learning algorithm, which also adopts a counting matrix as input. It isolates anomalies instead of profiling normal data points. It represents features as tree structures and anomalies are detected as instances with short average path lengths on the constructed isolation trees.
    \item OCSVM (One-Class Support Vector Machine) \cite{scholkopf2001estimating}: OCSVM is a variant of the Support Vector Machine algorithm that is designed for anomaly detection tasks \cite{li2003improving, wang2004anomaly}. The model is trained on normal data and finds the maximum margin hyperplane that separates the normal data from the origin.
    \item LogCluster \cite{lin2016log}: LogCluster is a density-based log clustering approach that groups similar log messages together. Anomalies are detected as log messages that do not belong to any cluster or belong to small clusters.
    \item DeepLog \cite{du2017deeplog}: DeepLog is a deep learning-based approach for anomaly detection in log data. It uses a long short-term memory (LSTM) network to model the log sequences and detect anomalies based on the prediction errors.
    \item LogAnomaly \cite{meng2019loganomaly}: LogAnomaly models a log stream as a natural language sequence, which can detect both sequential and quantitative log anomalies simultaneously. 

\end{itemize}
\begin{table*}[!h]
\centering
\caption{Experimental Results on HDFS, BGL, and Thunderbird Datasets.}
\label{tab:ad-res}
\resizebox{0.98\textwidth}{!}{%
\begin{tabular}{|c|ccc|ccc|ccc|}
\hline
\multirow{2}{*}{Method} & \multicolumn{3}{c|}{HDFS}                                                                           & \multicolumn{3}{c|}{BGL}                                                                            & \multicolumn{3}{c|}{Thunderbird}                                                                    \\ \cline{2-10} 
                        & \multicolumn{1}{c|}{Precision}         & \multicolumn{1}{c|}{Recall}            & F-1 score                & \multicolumn{1}{c|}{Precision}         & \multicolumn{1}{c|}{Recall}            & F-1 score                & \multicolumn{1}{c|}{Precision}         & \multicolumn{1}{c|}{Recall}            & F-1 score                \\ \hline
PCA                     & \multicolumn{1}{c|}{$0.166_{\pm0.008}$} & \multicolumn{1}{c|}{$0.059_{\pm0.003}$} & $0.087_{\pm0.002}$ & \multicolumn{1}{c|}{$0.117_{\pm0.023}$} & \multicolumn{1}{c|}{$0.035_{\pm0.007}$} & $0.054_{\pm0.010}$ & \multicolumn{1}{c|}{$0.953_{\pm0.004}$} & \multicolumn{1}{c|}{$0.980_{\pm0.005}$} & $0.966_{\pm0.003}$ \\ \hline
iForest                 & \multicolumn{1}{c|}{$0.043_{\pm0.010}$} & \multicolumn{1}{c|}{$0.422_{\pm0.224}$} & $0.078_{\pm0.021}$ & \multicolumn{1}{c|}{$0.491_{\pm0.364}$} & \multicolumn{1}{c|}{$0.037_{\pm0.052}$} & $0.063_{\pm0.090}$ & \multicolumn{1}{c|}{$0.338_{\pm0.128}$} & \multicolumn{1}{c|}{$0.015_{\pm0.011}$} & $0.028_{\pm0.020}$ \\ \hline
OCSVM                   & \multicolumn{1}{c|}{$0.058_{\pm0.012}$} & \multicolumn{1}{c|}{$0.910_{\pm0.089}$} & $0.108_{\pm0.021}$ & \multicolumn{1}{c|}{$0.073_{\pm0.003}$} & \multicolumn{1}{c|}{$0.345_{\pm0.010}$} & $0.121_{\pm0.004}$ & \multicolumn{1}{c|}{$0.550_{\pm0.004}$} & \multicolumn{1}{c|}{$0.998_{\pm0.000}$} & $0.709_{\pm0.003}$ \\ \hline
LogCluster              & \multicolumn{1}{c|}{$\bf{0.996_{\pm0.003}}$} & \multicolumn{1}{c|}{$0.368_{\pm0.001}$} & $0.538_{\pm0.001}$ & \multicolumn{1}{c|}{$\bf{0.941_{\pm0.015}}$} & \multicolumn{1}{c|}{$0.641_{\pm0.033}$} & $0.762_{\pm0.021}$ & \multicolumn{1}{c|}{$\bf{0.977_{\pm0.005}}$} & \multicolumn{1}{c|}{$0.291_{\pm0.063}$} & $0.445_{\pm0.067}$ \\ \hline
DeepLog                 & \multicolumn{1}{c|}{$0.793_{\pm0.092}$} & \multicolumn{1}{c|}{$0.863_{\pm0.031}$} & $0.824_{\pm0.060}$ & \multicolumn{1}{c|}{$0.792_{\pm0.048}$} & \multicolumn{1}{c|}{$0.946_{\pm0.012}$} & $0.861_{\pm0.028}$ & \multicolumn{1}{c|}{$0.864_{\pm0.005}$} & \multicolumn{1}{c|}{$0.997_{\pm0.000}$} & $0.926_{\pm0.003}$ \\ \hline
LogAnomaly              & \multicolumn{1}{c|}{$0.907_{\pm0.017}$} & \multicolumn{1}{c|}{$0.369_{\pm0.014}$} & $0.524_{\pm0.017}$ & \multicolumn{1}{c|}{$0.884_{\pm0.002}$} & \multicolumn{1}{c|}{$0.850_{\pm0.009}$} & $0.867_{\pm0.003}$ & \multicolumn{1}{c|}{$0.873_{\pm0.005}$} & \multicolumn{1}{c|}{$0.996_{\pm0.000}$} & $0.931_{\pm0.003}$ \\ \hline
OC4Seq              & \multicolumn{1}{c|}{$0.922_{\pm0.059}$} & \multicolumn{1}{c|}{$0.758_{\pm0.227}$} & $0.808_{\pm0.157}$ & \multicolumn{1}{c|}{$0.441_{\pm0.045}$} & \multicolumn{1}{c|}{$0.352_{\pm0.044}$} & $0.391_{\pm0.041}$ & \multicolumn{1}{c|}{$0.901_{\pm0.046}$} & \multicolumn{1}{c|}{$0.823_{\pm0.232}$} & $0.845_{\pm0.177}$ \\ \hline
LogBERT                 & \multicolumn{1}{c|}{$0.754_{\pm0.142}$} & \multicolumn{1}{c|}{$0.749_{\pm0.037}$} & $0.745_{\pm0.082}$ & \multicolumn{1}{c|}{$0.917_{\pm0.006}$} & \multicolumn{1}{c|}{$0.892_{\pm0.006}$} & $0.905_{\pm0.005}$ & \multicolumn{1}{c|}{$0.962_{\pm0.019}$} & \multicolumn{1}{c|}{$0.965_{\pm0.008}$} & $0.963_{\pm0.007}$ \\ \hline
CAT              & \multicolumn{1}{c|}{$0.102_{\pm0.022}$} & \multicolumn{1}{c|}{$0.422_{\pm0.082}$} & $0.164_{\pm0.034}$ & \multicolumn{1}{c|}{$0.177_{\pm0.122}$} & \multicolumn{1}{c|}{$0.210_{\pm0.184}$} & $0.190_{\pm0.148}$ & \multicolumn{1}{c|}{$0.751_{\pm0.072}$} & \multicolumn{1}{c|}{$0.516_{\pm0.124}$} & $0.607_{\pm0.120}$ \\ \hline\hline
LogGPT                  & \multicolumn{1}{c|}{$0.884_{\pm0.030}$} & \multicolumn{1}{c|}{$\bf{0.921_{\pm0.066}}$} & $\bf{0.901_{\pm0.036}^*}$ & \multicolumn{1}{c|}{$0.940_{\pm0.010}$} & \multicolumn{1}{c|}{$\bf{0.977_{\pm0.018}}$} & $\bf{0.958_{\pm0.011}^*}$ & \multicolumn{1}{c|}{$0.973_{\pm0.004}$} & \multicolumn{1}{c|}{$\bf{1.000_{\pm0.000}}$} & $\bf{0.986_{\pm0.002}^*}$ \\ \hline
\end{tabular}%
}
\\
\vspace{2pt}
\footnotesize{The asterisk indicates that LogGPT significantly outperforms the best baseline at the 0.05 level, according to the paired t-test.}
\end{table*}

\begin{table}[h]
\caption{Performance of LogGPT with or without reinforcement learning.}
\label{tab:ab-rl}
\centering
\resizebox{0.48\textwidth}{!}{%
\begin{tabular}{|c|c|c|c|c|}
\hline
Metric                     & Approach      & HDFS               & BGL                & Thunderbird        \\ \hline
\multirow{2}{*}{Precision} & LogGPT w/o RL & $0.932_{\pm0.015}$ & $0.936_{\pm0.011}$ & $0.971_{\pm0.004}$ \\ \cline{2-5} 
                           & LogGPT        & $0.884_{\pm0.030}$ & $0.940_{\pm0.010}$ & $0.973_{\pm0.004}$ \\ \hline
\multirow{2}{*}{Recall}    & LogGPT w/o RL & $0.790_{\pm0.101}$ & $0.975_{\pm0.018}$ & $1.000_{\pm0.000}$ \\ \cline{2-5} 
                           & LogGPT        & $0.921_{\pm0.066}$ & $0.977_{\pm0.018}$ & $1.000_{\pm0.000}$ \\ \hline
\multirow{2}{*}{F-1 score}        & LogGPT w/o RL & $0.853_{\pm0.065}$ & $0.955_{\pm0.010}$ & $0.985_{\pm0.002}$ \\ \cline{2-5} 
                           & LogGPT        & $0.901_{\pm0.036}^*$ & $0.958_{\pm0.011}$ & $0.986_{\pm0.002}^*$ \\ \hline
\end{tabular}
}
\\
\vspace{2pt}
\footnotesize{Significantly outperforms  LogGPT w/o RL at the 0.05 level (paired t-test).}
\end{table}

\begin{itemize}
    
    \item OC4Seq (Multi-Scale One-Class Recurrent Neural Networks) \cite{wang2021multi}: OC4Seq is designed to detect anomalies in discrete event sequences. 
    Recognizing that an anomalous sequence could be caused by individual events, subsequences of events, or the entire sequence, OC4Seq employs a multi-scale RNN framework to capture different levels of sequential patterns simultaneously.
    \item LogBERT \cite{guo2021logbert}: LogBERT is a BERT-based architecture to capture the patterns of normal log sequences via a log language model. LogBERT is trained to predict the masked log keys on normal log sequences and detects the abnormal log sequences based on the prediction errors. 
    \item CAT (Content-Aware Transformer) \cite{zhang2022cat}: CAT is a self-attentive encoder-decoder transformer framework designed for anomaly detection in event sequences. It incorporates the semantic information of event content by using a content-awareness layer to generate representations of each event. The encoder learns preamble event sequence representations with content awareness, and the decoder embeds sequences under detection into a latent space where anomalies are distinguishable.
\end{itemize}

\noindent\textbf{Implementation Details.}
We first employ Drain \cite{he2017drain} to parse raw log messages into log keys. For the baseline models, we utilize the Loglizer \cite{he2016experience} package to evaluate PCA, OCSVM, iForest, and LogCluster for anomaly detection. DeepLog and LogAnomaly are evaluated using the Deep-loglizer \cite{chen2021experience} package. For OC4Seq\footnote{https://github.com/KnowledgeDiscovery/OC4Seq}, LogBERT\footnote{https://github.com/HelenGuohx/logbert}, and CAT\footnote{https://github.com/mmichaelzhang/CAT}, we use the open-source code provided by the authors separately.

As for LogGPT, we use a GPT model with 6 layers and 6 heads. The dimensions of the embeddings and hidden states are set to 60. The learning rate is set to 1e-4 for the pre-training phase and 1e-6 for the fine-tuning phase. To accommodate different datasets, we set the K in Top-K to 50\% of the training log keys. It means during the test phase if an observed log key is not in the top 50\% of the prediction list from the GPT, the sequence will be labeled as an anomaly.
This allows us to maintain a high level of flexibility when dealing with datasets of varying sizes and characteristics. The batch size for the pre-training phase is set to 16, and we train the model for 100 epochs. The episode is set to 20 with early stop criteria to prevent overfitting and ensure efficient training. The code for LogGPT is publicly available\footnote{https://github.com/nokia/LogGPT}.

\subsection{Experimental Results}
\noindent\textbf{Performance on Log Anomaly Detection.} Table \ref{tab:ad-res} illustrates the results and standard deviation of LogGPT and various baselines over 10 runs on the HDFS, BGL, and Thunderbird datasets. The asterisk in the table indicates that LogGPT significantly outperforms the best baseline for each dataset at the 0.05 level, according to the paired t-test.

First, we can observe that PCA, iForest, and OCSVM perform poorly on the HDFS and BGL datasets, as indicated by their low F-1 scores. However, PCA's performance is notably better on the Thunderbird dataset, achieving a high F-1 score. This inconsistency in performance across datasets highlights the sensitivity of PCA to datasets.

LogCluster, specifically designed for log anomaly detection, shows improved performance over other traditional machine learning models, i.e., PCA, iForest, and OCSVM, on the HDFS and BGL datasets but is outperformed by PCA on the Thunderbird dataset. This pattern further emphasizes the importance of dataset-specific characteristics in determining the effectiveness of different methods.

Deep learning-based approaches, such as DeepLog, LogAnomaly, OC4seq, LogBERT, and CAT, outperform traditional methods across all three datasets, which shows the advantages of utilizing deep learning to capture complex patterns in log sequences. 

Our proposed model, LogGPT, stands out by consistently achieving the highest F-1 scores across all three datasets, with significant margins over all baselines.

\begin{figure*}[!h]
    \centering
    \begin{subfigure}[b]{0.32\textwidth}
        \centering
        \includegraphics[width=\textwidth]{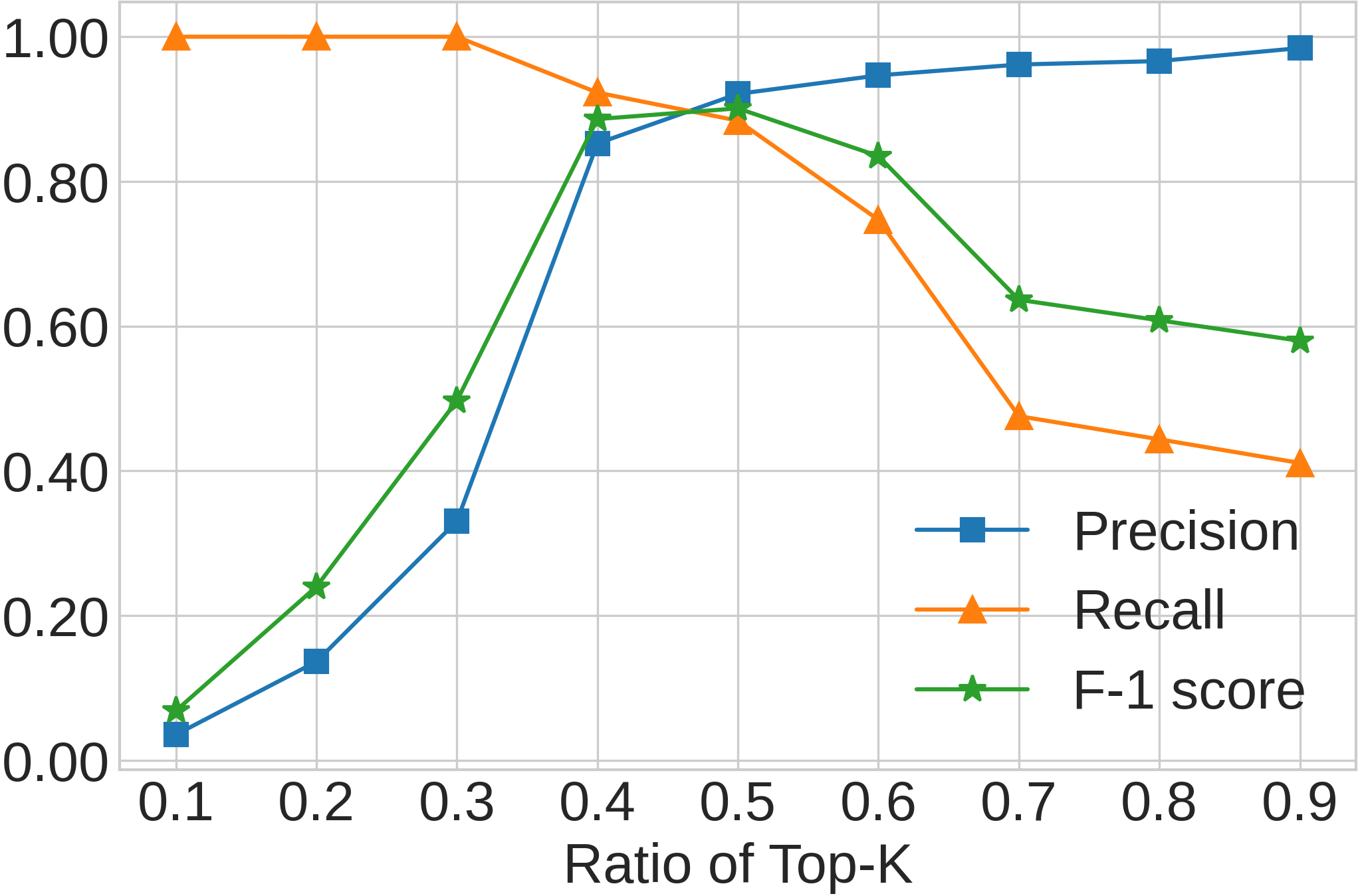}
        \caption{HDFS}
        \label{fig:ratio_HDFS}
    \end{subfigure}
    \begin{subfigure}[b]{0.32\textwidth}
        \centering
        \includegraphics[width=\textwidth]{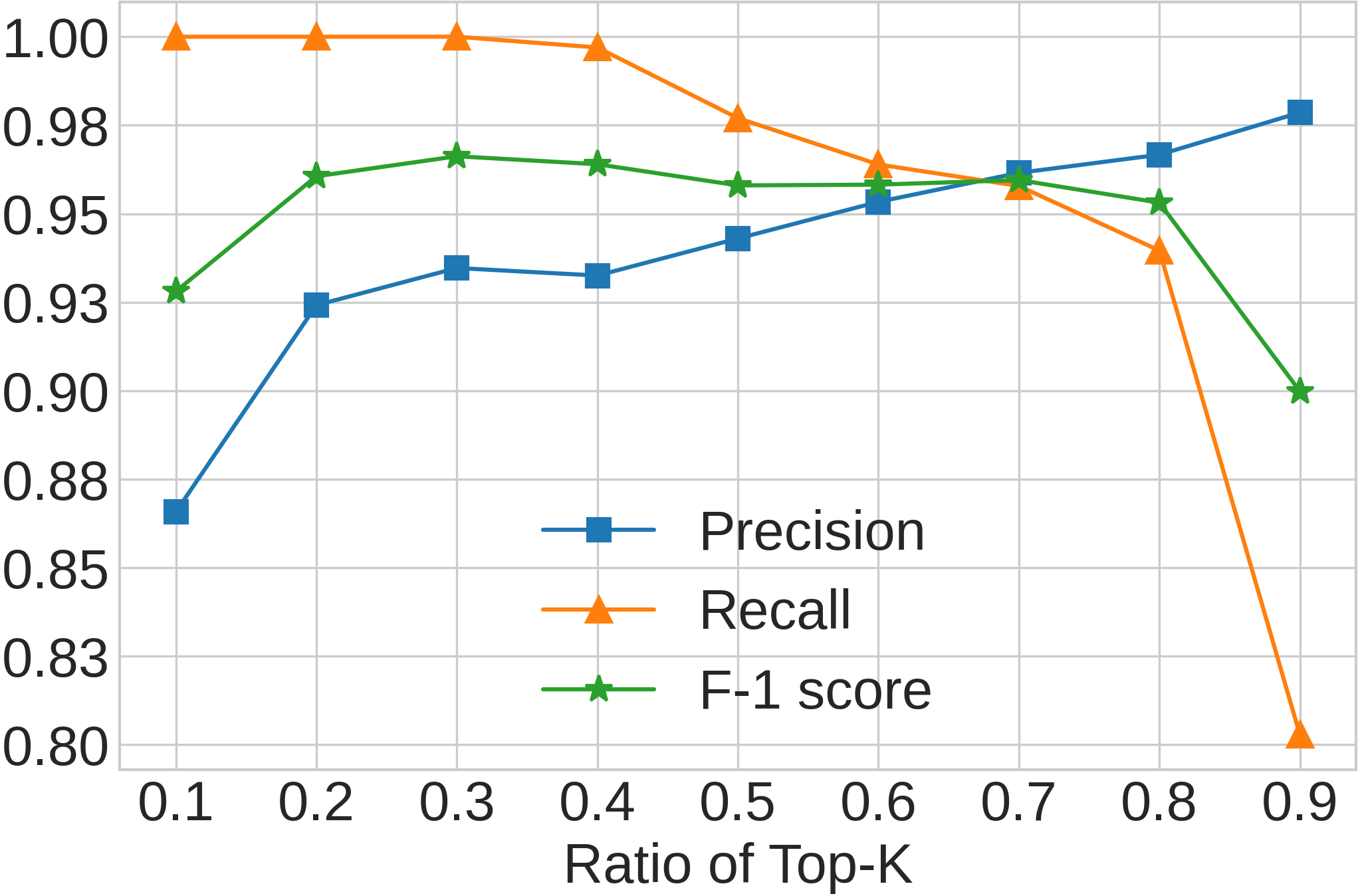}
        \caption{BGL}
        \label{fig:ratio_BGL}
    \end{subfigure}
    \begin{subfigure}[b]{0.32\textwidth}
        \centering
        \includegraphics[width=\textwidth]{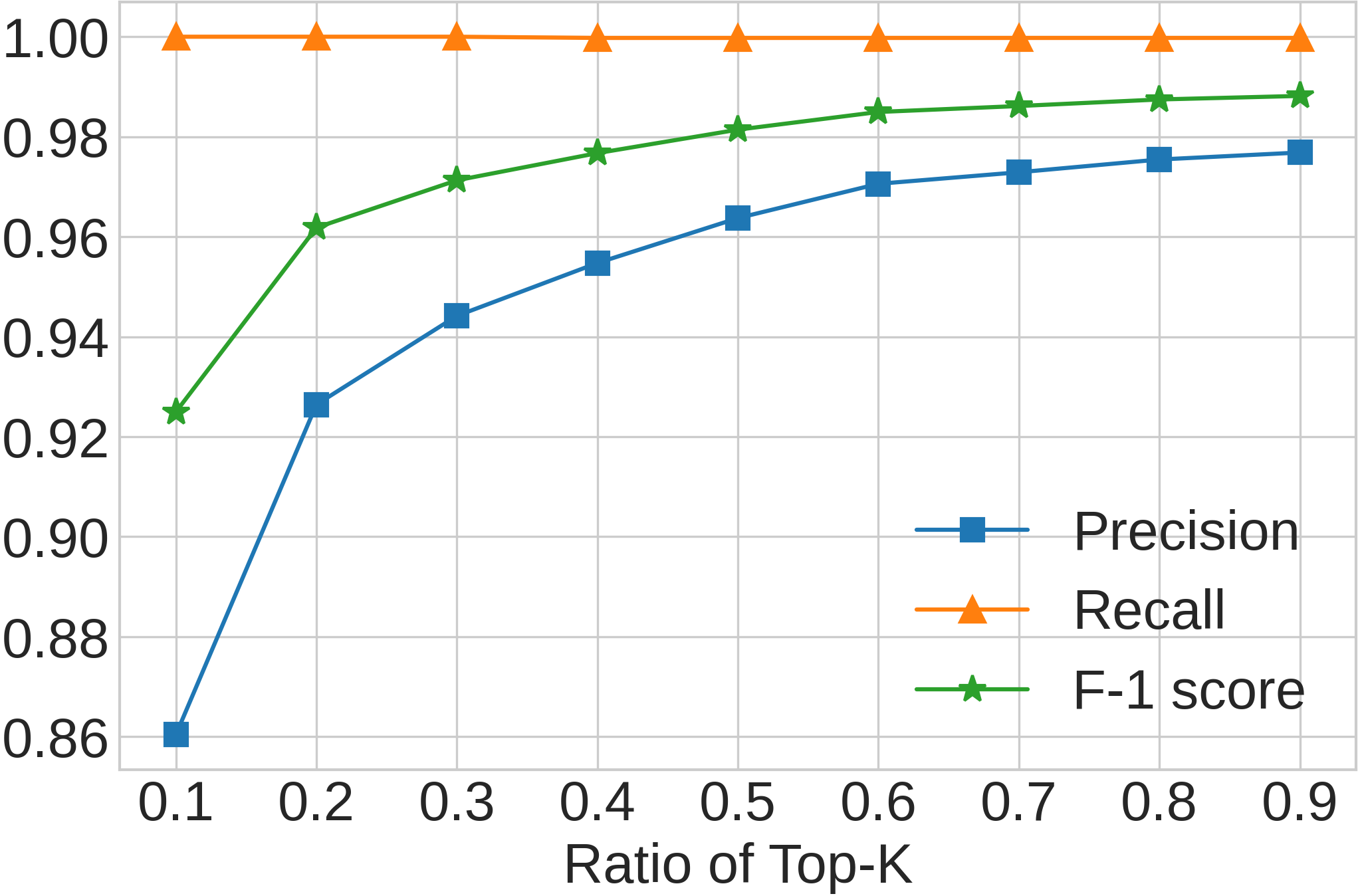}
        \caption{Thunderbird}
        \label{fig:ratio_TB}
    \end{subfigure}
    \caption{Impact of the ratio of Top-K log keys.}
    \label{fig:ratio}
\end{figure*}

\begin{figure*}[h]
    \begin{minipage}{\textwidth}
        \centering
        \includegraphics[width=0.8\textwidth]{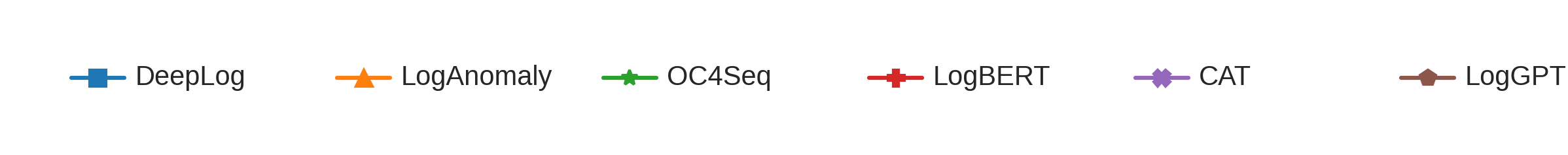}
    \end{minipage}

    \vspace{-0.5em} 
    \centering
    \begin{subfigure}{0.32\textwidth}
        \centering
        \includegraphics[width=\textwidth]{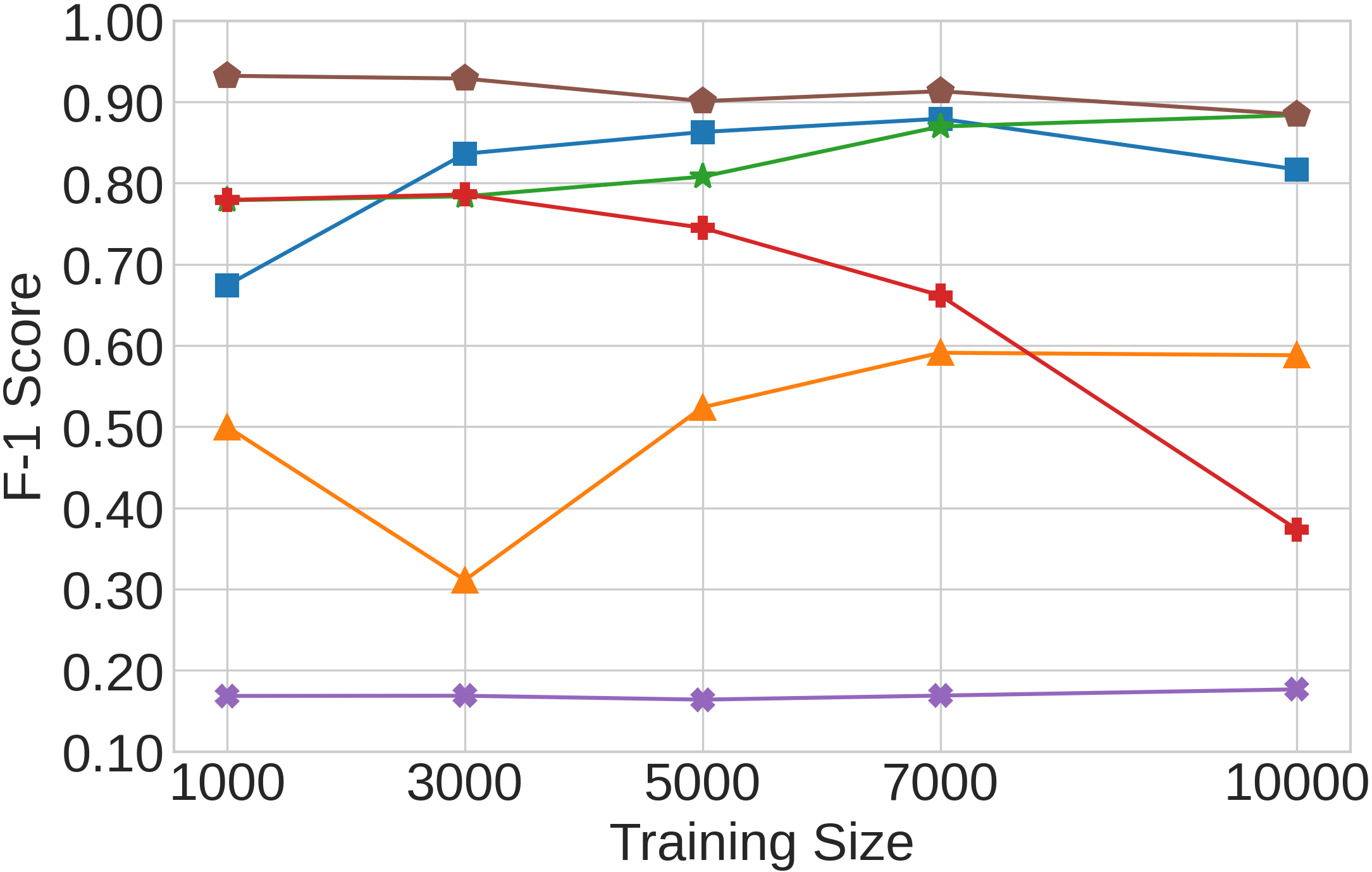}
        \caption{HDFS}
        \label{fig:hdfs_sample}
    \end{subfigure}%
    \begin{subfigure}{0.32\textwidth}
        \centering
        \includegraphics[width=\textwidth]{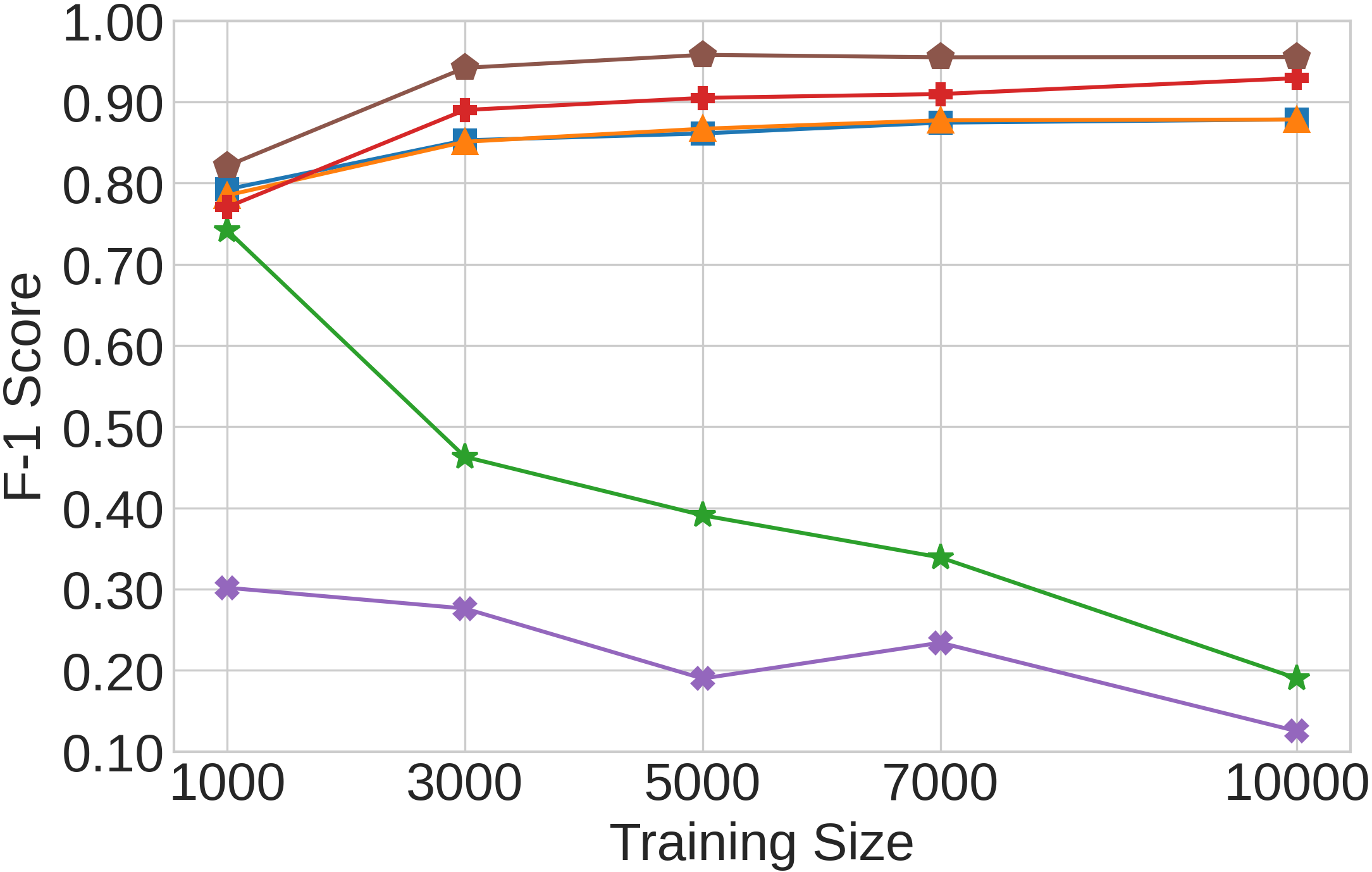}
        \caption{BGL}
        \label{fig:bgl_sample}
    \end{subfigure}%
    \begin{subfigure}{0.32\textwidth}
        \centering
        \includegraphics[width=\textwidth]{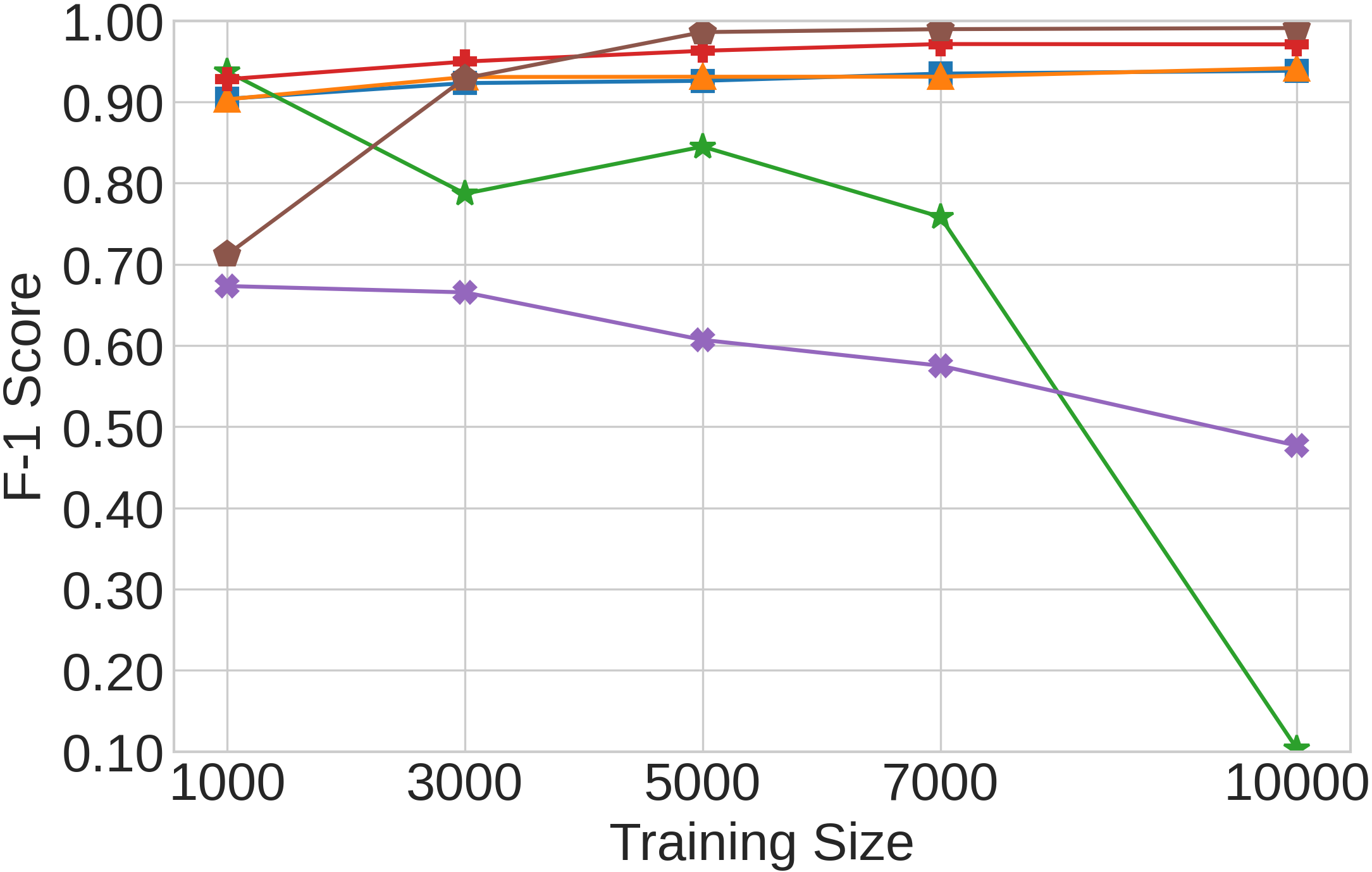}
        \caption{Thunderbird}
        \label{fig:tb_sample}
    \end{subfigure}
    \caption{Impact of the training size.}
    \label{fig:sample}
\end{figure*}

\noindent\textbf{Ablation Studies.} To investigate the contribution of reinforcement learning (RL) to the performance of LogGPT, we conducted an ablation study, comparing the performance of LogGPT with and without the RL component. The results are summarized in Table \ref{tab:ab-rl}.

First, we can notice that on both HDFS and Thunderbird datasets, LogGPT significantly outperforms LogGPT without the RL component, which demonstrates that the RL component enhances the overall performance of LogGPT for log anomaly detection. Especially, on the HDFS dataset, by finetuning the GPT model with RL reward, the recall achieved by LogGPT is improved with a large margin with a little sacrifice on precision, leading to extensive improvement in the F-1 score. It also shows that fine-tuning the log language model with Top-K reward can identify more log anomalies. Meanwhile, on the BGL dataset, we can also notice a slight improvement in F-1 of LogGPT compared to the one without the RL component. Another interesting finding is that even the LogGPT without the RL component already outperforms all baselines (shown in Table \ref{tab:ad-res}) in three datasets, which also shows the advantage of leveraging the GPT model to capture the patterns of log sequences.

\noindent\textbf{Parameter Analysis: Ratio of Top-K}. LogGPT detects the anomalies by examining whether the observed log key is in the Top-K list predicted by GPT. Therefore, K is an important parameter to determine the anomalies. We first analyze the difference in the performance by tuning K for anomaly detection. By default, K is set as 50\% of unique log keys. It means if the next log key falls into the top 50\% of unique log keys predicted by GPT, the sequence is normal.


The impact of different top-K ratios on the precision, recall, and F-1 score for the HDFS, BGL, and Thunderbird datasets is illustrated in Figure \ref{fig:ratio}. On both HDFS and BGL datasets, we have similar observations. With the increasing of ratios as normal log keys, the recall keeps decreasing when the ratio is greater than a threshold, such as 40\% in HDFS and BGL. This happens because when we have a large ratio, most of the keys are considered normal. In this case, the recall will be low. On the other hand, if the observed log key is predicted with an extremely low probability at a specific position, with a high chance, this log key is abnormal. Therefore, we can observe the increase in precision along with the increase in ratios.



For the Thunderbird dataset, the precision increases as the top-K ratio increases, while the recall remains almost constant, with a slight decrease at higher top-K ratios. The F-1 score increases steadily, reaching a peak at a specific top-K ratio. The reason for this behavior can be attributed to the inherent characteristics of the Thunderbird dataset. It is likely that the normal data within the Thunderbird dataset has high variability, which needs a broader range of acceptable continuations in the log sequences to reduce the false positive. As the top-K ratio increases, LogGPT becomes more selective in flagging anomalies, thereby increasing precision by reducing false positives. 

Overall, a low top-K ratio tends to lead to high recall but low precision, while a high top-K ratio leads to high precision but potentially lower recall. The optimal top-K ratio varies across datasets, reflecting the unique characteristics of each dataset.

\noindent\textbf{Scalability Analysis: Training Size.} It is well known that deep learning models usually require a sufficient number of training samples. The impact of training size on the performance of log anomaly detection models is critical. By analyzing the F-1 scores of various models across different training sizes, we can gain insights into their effectiveness and efficiency. In this experiment, we compare LogGPT with other deep learning-based baselines, across three datasets by varying the training size. Figure \ref{fig:sample} shows the experimental results.

The effect of the training size on the HDFS dataset reveals distinct patterns across different models (shown in Figure \ref{fig:hdfs_sample}). LogGPT demonstrates consistent performance across various training sizes, highlighting its robustness and ability to generalize well. OC4Seq shows a consistent increase in performance with the training size, indicating that it benefits from more extensive training data. DeepLog and LogAnomaly exhibit fluctuations in performance, which may be attributed to the sensitivity to training size. The decline in performance for LogBERT and stability for CAT may reflect limitations in their ability to leverage additional training data without changing other hyper-parameters. The varying behaviors of these models underscore the importance of carefully selecting the training size based on the model's characteristics.

We have similar observations on BGL and Thunderbird datasets. First, with larger training sizes, the performance of LogGPT, DeepLog, LogAnomaly, and LogBERT keep improving, which shows that these models can benefit from additional training data. Meanwhile, LogGPT can outperform those baselines in most cases. However, the sharp decline for OC4Seq and overall downward trend for CAT may indicate overfitting or challenges in generalizing from larger training sets.

Overall, LogGPT can achieve very good performance in three datasets. More training samples can further boost the performance of LogGPT.




\section{Conclusion}
In this work, we introduced LogGPT, a novel approach to log anomaly detection that builds upon GPT models, further enhanced by a reinforcement learning strategy. Through modeling log sequences as natural language, LogGPT innovatively adapts GPT for log anomaly detection. More importantly, recognizing the existing gap between language modeling and anomaly detection, LogGPT integrates a fine-tuning process guided by a novel Top-K reward metric for anomaly detection. Extensive experiments conducted across various datasets demonstrated the effectiveness of LogGPT, showcasing significant improvements over existing state-of-the-art methods.



\bibliographystyle{IEEEtran}
\bibliography{ref}

\begin{thebibliography}{10}
\providecommand{\url}[1]{#1}
\csname url@samestyle\endcsname
\providecommand{\newblock}{\relax}
\providecommand{\bibinfo}[2]{#2}
\providecommand{\BIBentrySTDinterwordspacing}{\spaceskip=0pt\relax}
\providecommand{\BIBentryALTinterwordstretchfactor}{4}
\providecommand{\BIBentryALTinterwordspacing}{\spaceskip=\fontdimen2\font plus
\BIBentryALTinterwordstretchfactor\fontdimen3\font minus
  \fontdimen4\font\relax}
\providecommand{\BIBforeignlanguage}[2]{{%
\expandafter\ifx\csname l@#1\endcsname\relax
\typeout{** WARNING: IEEEtran.bst: No hyphenation pattern has been}%
\typeout{** loaded for the language `#1'. Using the pattern for}%
\typeout{** the default language instead.}%
\else
\language=\csname l@#1\endcsname
\fi
#2}}
\providecommand{\BIBdecl}{\relax}
\BIBdecl

\bibitem{du2017deeplog}
M.~Du, F.~Li, G.~Zheng, and V.~Srikumar, ``Deeplog: Anomaly detection and
  diagnosis from system logs through deep learning,'' in \emph{Proceedings of
  the 2017 ACM SIGSAC conference on computer and communications security},
  2017, pp. 1285--1298.

\bibitem{guo2021logbert}
H.~Guo, S.~Yuan, and X.~Wu, ``Logbert: Log anomaly detection via bert,'' in
  \emph{2021 international joint conference on neural networks (IJCNN)}.\hskip
  1em plus 0.5em minus 0.4em\relax IEEE, 2021, pp. 1--8.

\bibitem{pang2021deep}
G.~Pang, C.~Shen, L.~Cao, and A.~V.~D. Hengel, ``Deep learning for anomaly
  detection: A review,'' \emph{ACM computing surveys (CSUR)}, vol.~54, no.~2,
  pp. 1--38, 2021.

\bibitem{le2022log}
V.-H. Le and H.~Zhang, ``Log-based anomaly detection with deep learning: How
  far are we?'' in \emph{Proceedings of the 44th international conference on
  software engineering}, 2022, pp. 1356--1367.

\bibitem{chalapathy2019deep}
R.~Chalapathy and S.~Chawla, ``Deep learning for anomaly detection: A survey,''
  \emph{arXiv preprint arXiv:1901.03407}, 2019.

\bibitem{landauer2023deep}
M.~Landauer, S.~Onder, F.~Skopik, and M.~Wurzenberger, ``Deep learning for
  anomaly detection in log data: A survey,'' \emph{Machine Learning with
  Applications}, vol.~12, p. 100470, 2023.

\bibitem{xu2009detecting}
W.~Xu, L.~Huang, A.~Fox, D.~Patterson, and M.~I. Jordan, ``Detecting
  large-scale system problems by mining console logs,'' in \emph{Proceedings of
  the ACM SIGOPS 22nd symposium on Operating systems principles}, 2009, pp.
  117--132.

\bibitem{liu2008isolation}
F.~T. Liu, K.~M. Ting, and Z.-H. Zhou, ``Isolation forest,'' in \emph{2008
  eighth ieee international conference on data mining}.\hskip 1em plus 0.5em
  minus 0.4em\relax IEEE, 2008, pp. 413--422.

\bibitem{wang2004anomaly}
Y.~Wang, J.~Wong, and A.~Miner, ``Anomaly intrusion detection using one class
  svm,'' in \emph{Proceedings from the Fifth Annual IEEE SMC Information
  Assurance Workshop, 2004.}\hskip 1em plus 0.5em minus 0.4em\relax IEEE, 2004,
  pp. 358--364.

\bibitem{meng2019loganomaly}
W.~Meng, Y.~Liu, Y.~Zhu, S.~Zhang, D.~Pei, Y.~Liu, Y.~Chen, R.~Zhang, S.~Tao,
  P.~Sun \emph{et~al.}, ``Loganomaly: Unsupervised detection of sequential and
  quantitative anomalies in unstructured logs.'' in \emph{IJCAI}, vol.~19,
  no.~7, 2019, pp. 4739--4745.

\bibitem{wang2021multi}
Z.~Wang, Z.~Chen, J.~Ni, H.~Liu, H.~Chen, and J.~Tang, ``Multi-scale one-class
  recurrent neural networks for discrete event sequence anomaly detection,'' in
  \emph{Proceedings of the 27th ACM SIGKDD conference on knowledge discovery \&
  data mining}, 2021, pp. 3726--3734.

\bibitem{vaswani2017attention}
A.~Vaswani, N.~Shazeer, N.~Parmar, J.~Uszkoreit, L.~Jones, A.~N. Gomez,
  {\L}.~Kaiser, and I.~Polosukhin, ``Attention is all you need,''
  \emph{Advances in neural information processing systems}, vol.~30, 2017.

\bibitem{devlin2018bert}
J.~Devlin, M.-W. Chang, K.~Lee, and K.~Toutanova, ``Bert: Pre-training of deep
  bidirectional transformers for language understanding,'' \emph{arXiv preprint
  arXiv:1810.04805}, 2018.

\bibitem{zhang2022cat}
S.~Zhang, Y.~Liu, X.~Zhang, W.~Cheng, H.~Chen, and H.~Xiong, ``Cat: Beyond
  efficient transformer for content-aware anomaly detection in event
  sequences,'' in \emph{Proceedings of the 28th ACM SIGKDD Conference on
  Knowledge Discovery and Data Mining}, 2022, pp. 4541--4550.

\bibitem{he2017drain}
P.~He, J.~Zhu, Z.~Zheng, and M.~R. Lyu, ``Drain: An online log parsing approach
  with fixed depth tree,'' in \emph{2017 IEEE international conference on web
  services (ICWS)}.\hskip 1em plus 0.5em minus 0.4em\relax IEEE, 2017, pp.
  33--40.

\bibitem{radford2019language}
A.~Radford, J.~Wu, R.~Child, D.~Luan, D.~Amodei, and I.~Sutskever, ``Language
  models are unsupervised multitask learners,'' 2019.

\bibitem{schulman2017proximal}
J.~Schulman, F.~Wolski, P.~Dhariwal, A.~Radford, and O.~Klimov, ``Proximal
  policy optimization algorithms,'' \emph{arXiv preprint arXiv:1707.06347},
  2017.

\bibitem{oliner2007supercomputers}
A.~Oliner and J.~Stearley, ``What supercomputers say: A study of five system
  logs,'' in \emph{37th annual IEEE/IFIP international conference on dependable
  systems and networks (DSN'07)}.\hskip 1em plus 0.5em minus 0.4em\relax IEEE,
  2007, pp. 575--584.

\bibitem{xu2009largescale}
W.~Xu, L.~Huang, A.~Fox, D.~Patterson, and M.~Jordan, ``Largescale system
  problem detection by mining console logs,'' \emph{Proceedings of SOSP’09},
  2009.

\bibitem{scholkopf2001estimating}
B.~Sch{\"o}lkopf, J.~C. Platt, J.~Shawe-Taylor, A.~J. Smola, and R.~C.
  Williamson, ``Estimating the support of a high-dimensional distribution,''
  \emph{Neural computation}, vol.~13, no.~7, pp. 1443--1471, 2001.

\bibitem{li2003improving}
K.-L. Li, H.-K. Huang, S.-F. Tian, and W.~Xu, ``Improving one-class svm for
  anomaly detection,'' in \emph{Proceedings of the 2003 international
  conference on machine learning and cybernetics (IEEE Cat. No. 03EX693)},
  vol.~5.\hskip 1em plus 0.5em minus 0.4em\relax IEEE, 2003, pp. 3077--3081.

\bibitem{lin2016log}
Q.~Lin, H.~Zhang, J.-G. Lou, Y.~Zhang, and X.~Chen, ``Log clustering based
  problem identification for online service systems,'' in \emph{Proceedings of
  the 38th International Conference on Software Engineering Companion}, 2016,
  pp. 102--111.

\bibitem{he2016experience}
S.~He, J.~Zhu, P.~He, and M.~R. Lyu, ``Experience report: System log analysis
  for anomaly detection,'' in \emph{2016 IEEE 27th international symposium on
  software reliability engineering (ISSRE)}.\hskip 1em plus 0.5em minus
  0.4em\relax IEEE, 2016, pp. 207--218.

\bibitem{chen2021experience}
Z.~Chen, J.~Liu, W.~Gu, Y.~Su, and M.~R. Lyu, ``Experience report: Deep
  learning-based system log analysis for anomaly detection,'' \emph{arXiv
  preprint arXiv:2107.05908}, 2021.

\end{thebibliography}

\end{document}